\documentclass[letterpaper,conference]{IEEEtran}
\usepackage{amsmath,amsfonts}
\usepackage{algorithmic}
\usepackage{algorithm}
\usepackage{array}
\usepackage[caption=false,font=normalsize,labelfont=sf,textfont=sf]{subfig}
\usepackage{textcomp}
\usepackage{stfloats}
\usepackage{url}
\usepackage{verbatim}
\usepackage{graphicx}
\usepackage{cite}
\usepackage{tabularx}

\renewcommand{\baselinestretch}{0.925}
\newlength{\bibitemsep}
\newlength{\bibparskip}
\let\oldthebibliography\thebibliography
\renewcommand\thebibliography[1]{%
  \oldthebibliography{#1}%
  \setlength{\parskip}{\bibitemsep}%
  \setlength{\itemsep}{\bibparskip}%
}

\makeatletter
\def\ps@headings{\def\@oddhead{\IEEEdoarxivheader{-1\oddsidemargin}\relax
\hbox{}\@IEEEheaderstyle\rightmark\hfil\thepage}\relax
\def\@evenhead{\IEEEdoarxivheader{-1\evensidemargin}\relax
\hbox{}\@IEEEheaderstyle\rightmark\hfil\thepage}\relax
\def\@oddfoot{\IEEEdoarxivfooter{-1\oddsidemargin}\hfil\hbox{}}\relax
\def\@evenfoot{\IEEEdoarxivfooter{-1\evensidemargin}\hfil\hbox{}}\relax}
\def\ps@IEEEtitlepagestyle{\ps@headings}

\def\IEEEarxivheadfootoffset{3pt}
\newdimen\IEEEheadtotopofpage
\newdimen\IEEEfoottobottomofpage
\newbox\@IEEEboxX

\def\IEEEarxivheader{}
\def\IEEEarxivfooter{}

\advance\IEEEheadtotopofpage by 1\topmargin
\advance\IEEEheadtotopofpage by 1in\relax

\IEEEfoottobottomofpage\paperheight
\advance\IEEEfoottobottomofpage by -1in\relax
\advance\IEEEfoottobottomofpage by -1\topmargin
\advance\IEEEfoottobottomofpage by -1\headheight
\advance\IEEEfoottobottomofpage by -1\headsep
\advance\IEEEfoottobottomofpage by -1\textheight
\advance\IEEEfoottobottomofpage by -1\footskip
\def\IEEEarxivheaderstyle{\normalfont\footnotesize}

\def\IEEEdoarxivheader#1{\@IEEEtrantmpdimenA\IEEEarxivheadfootoffset\relax
\@IEEEtrantmpdimenA -1\@IEEEtrantmpdimenA
\advance\@IEEEtrantmpdimenA by \IEEEheadtotopofpage
\settoheight{\@IEEEtrantmpdimenB}{\IEEEarxivheaderstyle HT}\relax
\advance\@IEEEtrantmpdimenA by -1\@IEEEtrantmpdimenB
\setbox\@IEEEboxX=\hbox{\relax
\raisebox{\@IEEEtrantmpdimenA}[0pt][0pt]{\parbox[t]{\textwidth}{\centering
\IEEEarxivheaderstyle\IEEEarxivheader}}}\relax
\wd\@IEEEboxX=0pt\relax
\ht\@IEEEboxX=0pt\relax
\dp\@IEEEboxX=0pt\relax
\box\@IEEEboxX\relax}

\def\IEEEarxivfooterstyle{\normalfont\footnotesize}

\def\IEEEdoarxivfooter#1{\@IEEEtrantmpdimenA\IEEEfoottobottomofpage\relax
\advance\@IEEEtrantmpdimenA by \IEEEarxivheadfootoffset\relax
\settodepth{\@IEEEtrantmpdimenB}{\IEEEarxivheaderstyle gjpqy}\relax
\advance\@IEEEtrantmpdimenA by 1\@IEEEtrantmpdimenB
\setbox\@IEEEboxX=\hbox{\hskip#1\hskip -1in\relax
\raisebox{\@IEEEtrantmpdimenA}[0pt][0pt]{\parbox[b]{\paperwidth}{\centering
\IEEEarxivfooterstyle\IEEEarxivfooter}}}\relax
\wd\@IEEEboxX=0pt\relax
\ht\@IEEEboxX=0pt\relax
\dp\@IEEEboxX=0pt\relax
\box\@IEEEboxX\relax}

\def\@IEEEheaderstyle{\normalfont\scriptsize}
\def\@IEEEfooterstyle{\normalfont\scriptsize}

\makeatother

\renewcommand{\IEEEarxivheadfootoffset}{3pt}
\renewcommand{\IEEEarxivheaderstyle}{\normalfont\footnotesize}
\renewcommand{\IEEEarxivfooterstyle}{\normalfont\scriptsize}
\renewcommand{\IEEEarxivheader}{This is the accepted version of an article
that has been published in 2022 IEEE/RSJ International Conference on Intelligent Robots and Systems (IROS). Changes were made to this version
by the publisher prior to publication.\\
The final version of record is available at 
\url{https://doi.org/10.1109/IROS47612.2022.9982247}}
\renewcommand{\IEEEarxivfooter}{ %
© 2022 IEEE.  Personal use of this material is permitted.  Permission from IEEE must be obtained for all other uses, in any current or future media, including reprinting/republishing this   material \\ for advertising or promotional purposes, creating new collective works, for resale or redistribution to servers or lists, or reuse of any copyrighted component of this work in other works}

\makeatother          

\begin{document}

\title{\LARGE \bf Cutaneous Feedback Interface for Teleoperated In-Hand Manipulation}

\author{Yaonan Zhu, Jacinto Colan, Tadayoshi Aoyama, and Yasuhisa Hasegawa
\thanks{Yaonan Zhu and Jacinto Colan are co-first authors. (Corresponding author: Jacinto Colan)}
\thanks{The authors are with Department of Micro-Nano Mechanical Science and Engineering, Nagoya University, Nagoya, Japan. \{zhu, colan\}@robo.mein.nagoya-u.ac.jp, \{aoyama, hasegawa\}@mein.nagoya-u.ac.jp}
\thanks{This work is supported by JST AIP Grant Number JPMJCR20G8, Japan, JST CREST Grant Number JPMJCR20D5, Japan.
}}

\maketitle

\begin{abstract}
In-hand pivoting is one of the important manipulation skills that leverage robot grippers' extrinsic dexterity to perform repositioning tasks to compensate for environmental uncertainties and imprecise motion execution.
Although many researchers have been trying to solve pivoting problems using mathematical modeling or learning-based approaches, the problems remain as open challenges.
On the other hand, humans perform in-hand manipulation with remarkable precision and speed.
Hence, the solution could be provided by making full use of this intrinsic human skill through dexterous teleoperation. 
For dexterous teleoperation to be successful, 
interfaces that enhance and complement haptic feedback are of great necessity.  
In this paper, we propose a cutaneous feedback interface that complements the somatosensory information humans rely on when performing dexterous skills. The interface is designed based on five-bar link mechanisms and provides two contact points in the index finger and thumb for cutaneous feedback. By integrating the interface with a commercially available haptic device, the system can display information such as grasping force, shear force, friction, and grasped object's pose. Passive pivoting tasks inside a numerical simulator Isaac Sim is conducted to evaluate the effect of the proposed cutaneous feedback interface. 
\end{abstract}

\section{Introduction}
\IEEEPARstart{I}{n-hand} manipulation is an intrinsic capability of human hands. Through precisely controlled synergies of finger motions, humans can re-position, rotate, slide and push grasped objects for complex manipulation tasks \cite{karayiannidis2016adaptive}. This sophisticated skill has fascinated robotics researchers for decades, which forms an important research field widely known as \textit{dexterous manipulation} \cite{salisbury1982articulated,okamura2000overview}.
Generally, in-hand manipulation can be categorised into two subfields. One is to study intrinsic dexterity, and the other one is to study extrinsic dexterity. Many researches have been focused on the former intrinsic dexterity with multi-fingered robotic grippers, and sophisticated learning based algorithms \cite{andrychowicz2020learning, li2019vision, tanikawa1999development}. 

However, in most of the cases, robotic platforms are equipped with robust and cost-efficient parallel jaw grippers to simplify the control and grasping planning process \cite{zhu20216}. Although, the lack of mechanical complexity and actuation lead to the degraded dexterity, it can be compensated by adopting extrinsic dexterity. Here, extrinsic dexterity is to make full use of resources external to robot grippers such as gravity, inertial forces, and external contacts. \cite{dafle2014extrinsic}.     

In this paper, we focus on in-hand pivoting, which is one of the important in-hand manipulation skills that uses this extrinsic dexterity to reposition objects, and it is greatly demanded in robotic applications such as, motion planning \cite{9391990}, tool manipulation \cite{Hou-2018-105030}, or surgical applications\cite{colan21}. 
In most of the robotic tasks, the ability of grasping objects in the correct pose with respect to the gripper coordinate is required. 
Although, the planning algorithms could plan the robot motion correctly, the resulting object pose may be not the desired one due to environment uncertainties and imprecise motion execution \cite{8202299}. As a consequence, to successfully execute the desired tasks, robots need to reposition the items within their gripper. By adopting the in-hand manipulation, the re-positioning process can be achieved effectively without using additional space for regrasping.

Many researches have been trying to solve pivoting problems by using analytical modeling methods to plan adequate grasping torques or generate certain inertia \cite{hou2018fast,6907062}. Successful pivoting relies on precisely modeled gravity and friction parameters during the interaction process between gripper and objects, or with external environments. In the conventional approaches, how to precisely estimate those critical parameters become an important issue. This friction estimation can be more crucial when considering the scenario where robots need to handle novel objects. Introducing a closed-loop control could be a help to the situation. In \cite{karayiannidis2016adaptive}, Karayiannidis et.al. implemented a closed loop adaptive controller for in-hand pivoting tasks by integrating tactile and visual-based feedback. The work successfully performed passive pivoting tasks under uncertainties.

On the other hand, humans can perform this dexterous motion with remarkable precision. Hence, there could be another approach to solve in-hand pivoting by leveraging this instinctive human skill. 
Robot teleoperation has been widely used as a bridge between human and machines since it was first invented by Goertz in 1940s \cite{goertz1964manipulator}. Through carefully designed teleoperation systems, human can control robots to perform human-like motions and complete tasks otherwise not possible without human input \cite{hokayem2006bilateral}. Moreover, in recent years, teleoperation systems have been widely applied to learning based manipulations that robot learns human manipulation skills \cite{losey2021learning,delpreto2020helping}.

To successfully transfer human dexterity to robots and achieve high quality in-hand manipulation, interfaces that provides adequate haptic feedback are of great necessity \cite{wildenbeest2012impact, zhu2020enhancing}. Although, most of the researches focus on enhancing the force feedback, to perform dexterous telemanipulation, additional haptic sensation feedback to inform friction on finger tips is important.
In this paper we propose a cutaneous feedback interface that can inform users the friction and object poses when performing passive pivoting. 

Contributions of this paper are summarised as follows:
\begin{itemize}
  \item This paper proposes a cutaneous feedback device based on multi-point contact five-bar mechanism for real-time in-hand object pose estimation.
  \item The basic performance of the device is analysed and verified by conducting a skin deformation pattern recognition experiment.
  \item This paper modelled objects’ force interaction inside the latest numerical simulator Nvidia Isaac Sim, and conducted teleoperated in-hand passive pivoting experiments to verify the cutaneous feedback effects on human subjects.
\end{itemize}

\begin{figure}[t]
	\centering
	\includegraphics[width=2.5in]{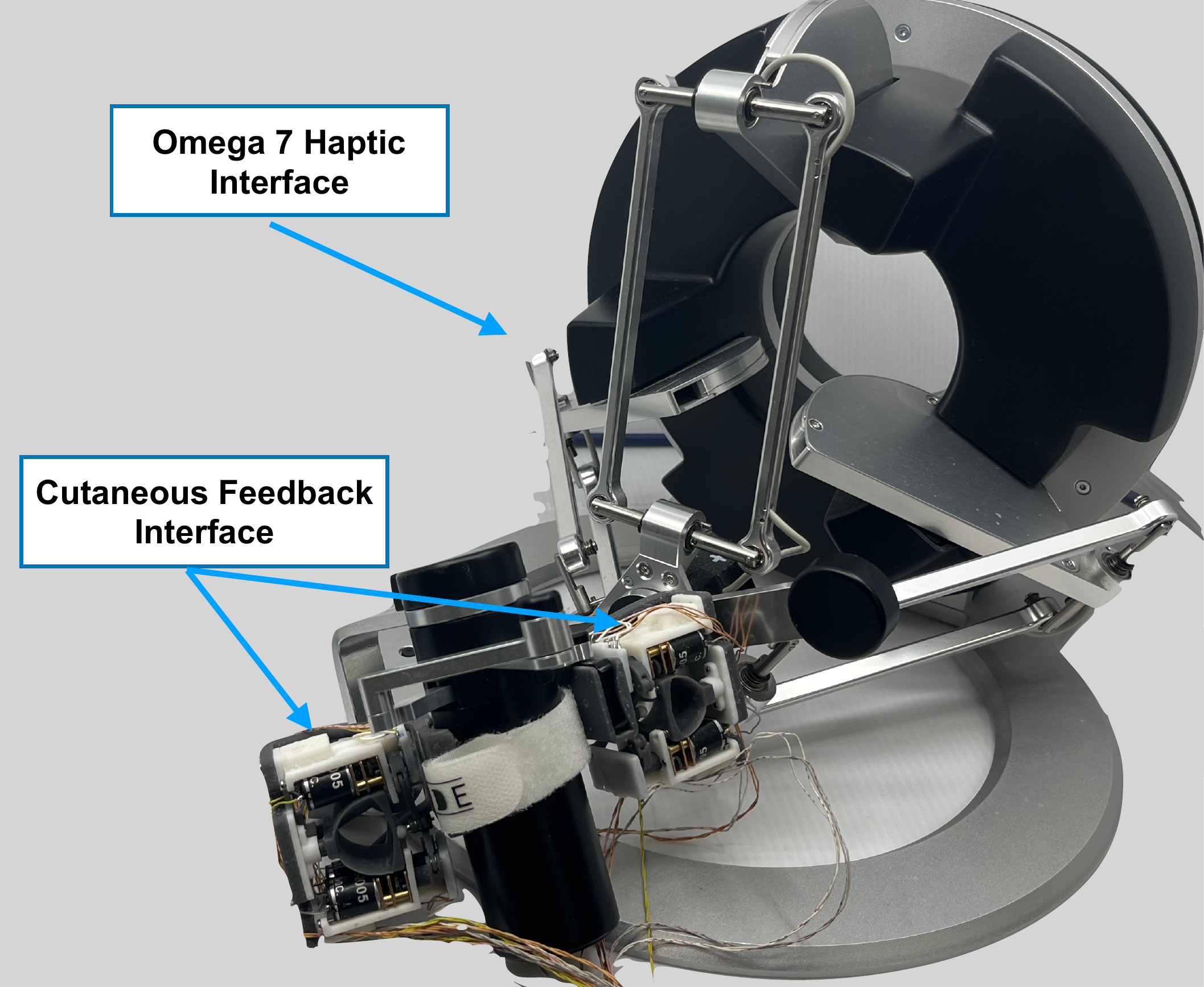}
	\caption{Cutaneous feedback interface attatched on Omega 7 haptic device}
	\label{fig:1}
\end{figure}

\section{Related Works}

Tactile haptic devices relies on shear forces applied to the user's fingertip, to resemble the skin deformation produced when interacting with an object. The increased sensitivity and directional cues of shear forces provides additional information over the use of normal forces \cite{culbertson18}. Several wearable tactile haptic devices have been proposed in previous works \cite{wang20}. Overall, they can be categorized in: belt/wire based, tactor-displacement based and linkage based devices\cite{girard16}.

Belt-based fingertip haptic devices have been proposed for rendering shear forces  that integrate tactile, force and thermal feedback \cite{murakami17}. Tactor-displacement based haptic devices has also been proposed to render shear forces through the displacement of tactors with the fingertip \cite{girard16}. Link-based haptic devices are attractive because of their compactness and versatility. Leonardis et al. \cite{leonardis15} proposed a wearable haptic device for 3D fingertip skin stretch rendering by a parallel mechanism comprising three revolute-spherical-revolute kinematic chains. A similar approach was followed in \cite{prattichizzo13} and \cite{chinello18}. Prattichizzo et al. proposed a 3-DoF wearable device based on a mobile platform actuated by wires to apply forces in different directions to the fingertip. Chinello et al. developed a wearable fingertip device based on three revolute-revolute-spherical kinematic chains to provide normal forces in different orientations that simulate contacts with arbitrarily oriented surfaces. Single five-bar linkage mechanisms have been proposed for haptic feedback in \cite{ballardini18},\cite{tsetserukou14}. Tsetserukou et al. \cite{tsetserukou14} proposed a 2-DoF cutaneous fingertip interface based on a five-bar linkage mechanism to generate normal forces along the mechanism axis. In \cite{ballardini18}, the five-bar mechanism is proposed to apply tactile stimulations to the palm. 

In this work, we explored the design of a novel tactile haptic device for skin deformation through multiple contact points in contrast to the common single contact point approach.  

\section{Cutaneous Haptic Feedback Device}
\subsection{Device Description}

\begin{figure}[t]
	\centering
	\includegraphics[width=3in]{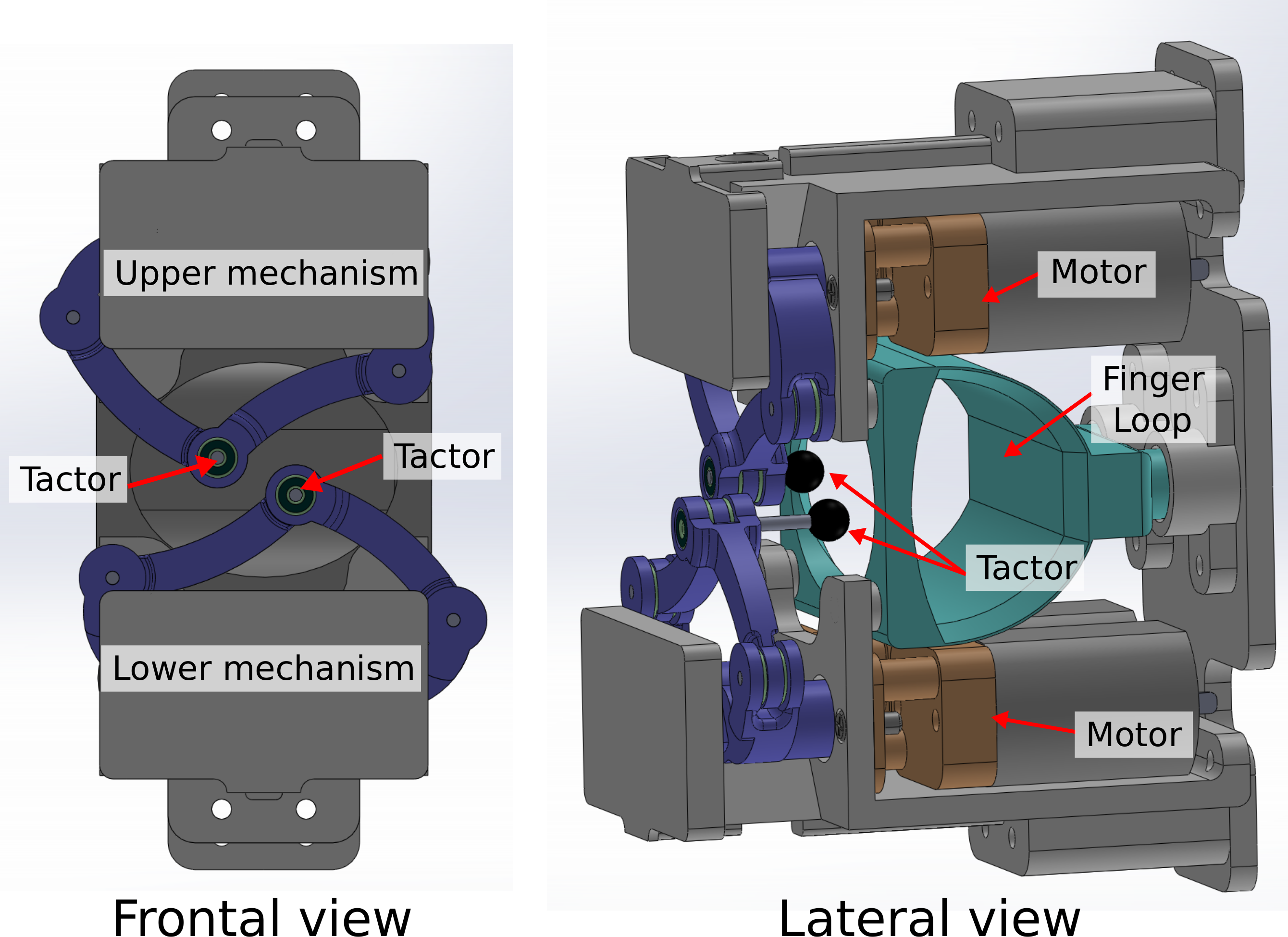}
	\caption{ A CAD rendering of a single tactile device.}
	\label{fig:2}
\end{figure}

The device is composed of two tactile device stations, one for the index finger and one for the thumb. As Fig.~\ref{fig:1} shows, the device is attached to a commercially available force feedback device (Omega7, Force Dimension). Each station comprises two five-bar linkage mechanism sharing a fingertip workspace as shown in Fig.~\ref{fig:2}. Each five-bar mechanism is driven by two DC motors 752-2005 (RS PRO). The motor rotation angles are obtained from precision potentiometers installed along the axis of each motor. The base and links of each device station were custom printed on a Formlabs 2 3D printer using a Tough2000 resin (Formlabs Co.). Table~\ref{tab:1} includes the origin coordinates of the lower ($O_1$, $O_2$) and upper ($O_3$, $O_4$) base links and link dimensions ($L_1$, $L_2$, $L_3$, $L_4$) for both finger.

\begin{table}[!t]
\caption{Mechanism dimensions (mm)}

\centering
\begin{tabular}{ |c|c|c|  }

 \hline
  \bf{Finger}   & $\bf{Index}$    & $\bf{Thumb}$ \\
 \hline
 $\bf{O_1}$    & (0,0)    & (0,0)   \\
 $\bf{O_2}$   & (12.5,0)    & (12.5,0)  \\
 $\bf{O_3}$    & (0,31)    & (0,35)   \\
 $\bf{O_4}$   & (12.5,31)    & (12.5,35)  \\
 $\bf{L_1}$    & 9    & 9   \\
 $\bf{L_2}$   & 9    & 9  \\
 $\bf{L_3}$    & 15    & 17.5   \\
 $\bf{L_4}$   & 15    & 17.5  \\
 \hline
\end{tabular}

\label{tab:1}
\end{table}

\subsection{Kinematic Analysis}
In \cite{bourbonnais15}, a geometry based direct and inverse kinematics for different linkage configurations is described. In the following subsection we summarize the direct and inverse kinematic analysis used for the proposed device.

\begin{figure}[t]
	\centering
	\includegraphics[width=3in]{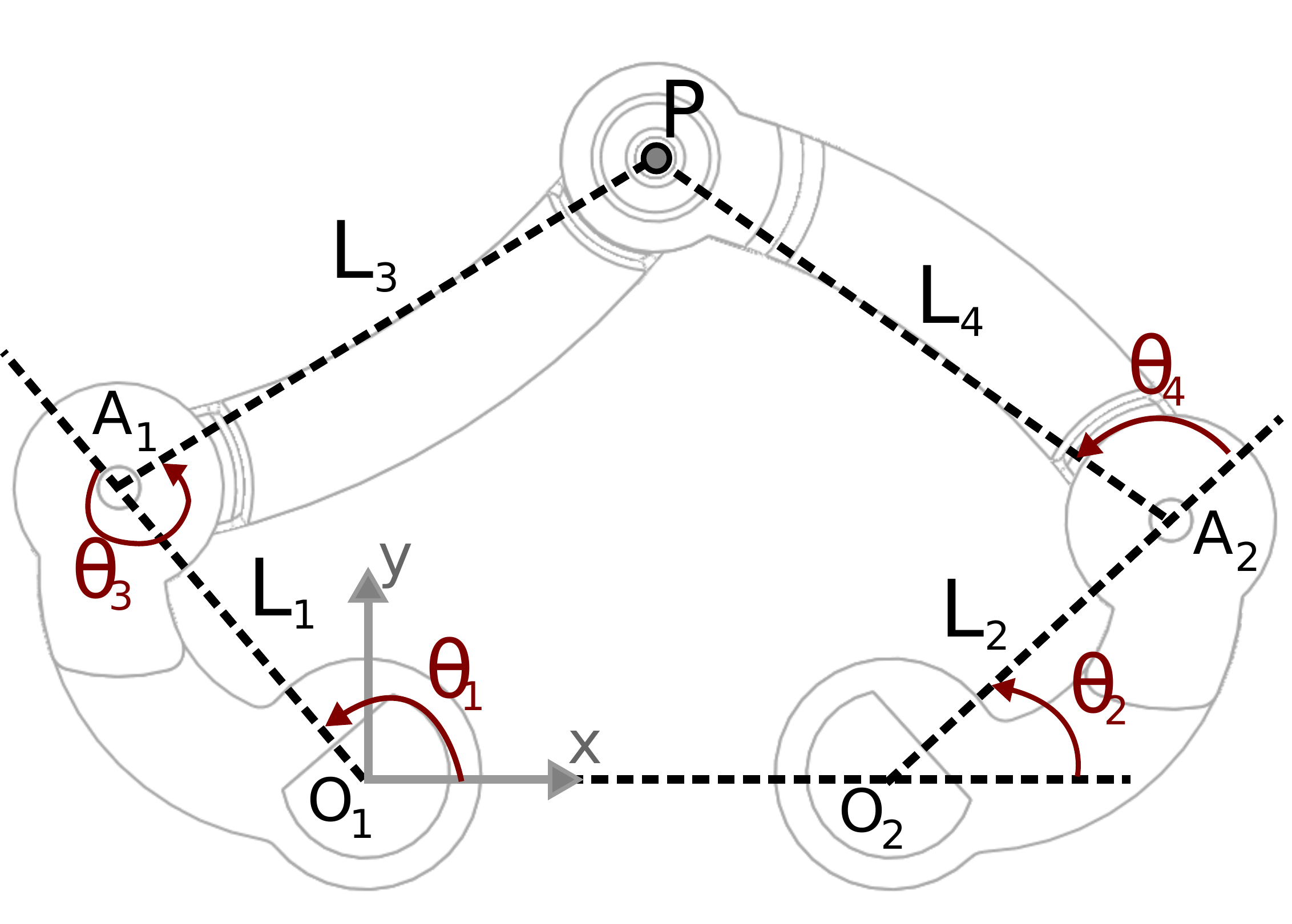}
	\caption{ Link diagram of the lower five-bar link mechanism }
	\label{fig:3}
\end{figure}

For simplification, we consider only the analysis for the lower five-bar linkage mechanism. The same algorithm works for the upper mechanism. Each mechanism comprises four links ($L_1, L_2, L_3, L_4$) and two degrees of freedom are controlled ($\theta_1, \theta_2$) as shown in Fig.~\ref{fig:3}.

\subsubsection{Direct Kinematics}
The Cartesian position of the intermediate joints  $\mathbf{r}_{A_1}$, $\mathbf{r}_{A_2}$ can be obtained from the controllable joints $\theta_1$,$\theta_2$ by:
\begin{equation}
	\begin{aligned}
        \mathbf{r}_{A_{1,2}} = \mathbf{r}_{O_{1,2}} + L_{1,2}
        \begin{bmatrix} 
        \cos{\theta_{1,2}} \\ 
        \sin{\theta_{1,2}} 
        \end{bmatrix}
	\end{aligned}
	\label{eq:1}
\end{equation}

The position of the tactor $\mathbf{r}_P$ can then be obtained from the following direct kinematics equation:

\begin{equation}
	\begin{aligned}
        \mathbf{r}_{P} = \mathbf{r}_{A_{1,2}} + \frac{L_{3,4}}{\|\mathbf{r}_{A_{2}} - \mathbf{r}_{A_{1}}\| }       \begin{bmatrix} 
                \cos{\alpha}  & \mp \sin{\alpha} \\ 
                \pm \sin{\alpha} & \cos{\alpha} 
            \end{bmatrix}
            \left( \mathbf{r}_{A_{2}} - \mathbf{r}_{A_{1}} \right)
	\end{aligned}
	\label{eq:2}
\end{equation}

with $\alpha = \arccos{ \left( \frac{L_3^2-L_4^2+\| \mathbf{r_{A_{2}} - r_{A_{1}}}\|^2}{2L_3 \|\mathbf{r}_{A_{2}} - \mathbf{r}_{A_{1}}\|} \right) }$.

\subsubsection{Inverse  Kinematics}
The control angles ($\theta_1$, $\theta_2$) can be obtained from a desired tactor position $\mathbf{r}_P$ by:

\begin{equation}
	\begin{aligned}
	    \theta_{1,2} = \arctan{\left( {r_y}_{A_{1,2}} - {r_y}_{O_{1,2}} , {r_x}_{A_{1,2}} - {r_x}_{O_{1,2}} \right)}
	\end{aligned}
	\label{eq:3}
\end{equation}
Where the Cartesian positions of the intermediate joints are computed as:

\begin{equation}
	\begin{aligned}
        \mathbf{r}_{A_i} = \mathbf{r}_{O_i} + \frac{L_i}{\|\mathbf{r}_{P} - \mathbf{r}_{O_i}\| }       \begin{bmatrix} 
                \cos{\alpha_i}  & \mp \sin{\alpha_i} \\ 
                \pm \sin{\alpha_i} & \cos{\alpha_i} 
            \end{bmatrix}
            \left( \mathbf{r}_{P} - \mathbf{r}_{O_i} \right)
	\end{aligned}
	\label{eq:4}
\end{equation}

with $\alpha_i = \arccos{ \left( \frac{L_i^2 - L_{i+2}^2+\| \mathbf{r_{P} - r_{O_{i}}}\|^2}{2L_i \|\mathbf{r}_{P} - \mathbf{r}_{O_{i}}\|} \right) }$.

\subsubsection{Reachable workspace}
\begin{figure}[t]
	\centering
	\includegraphics[width=3in]{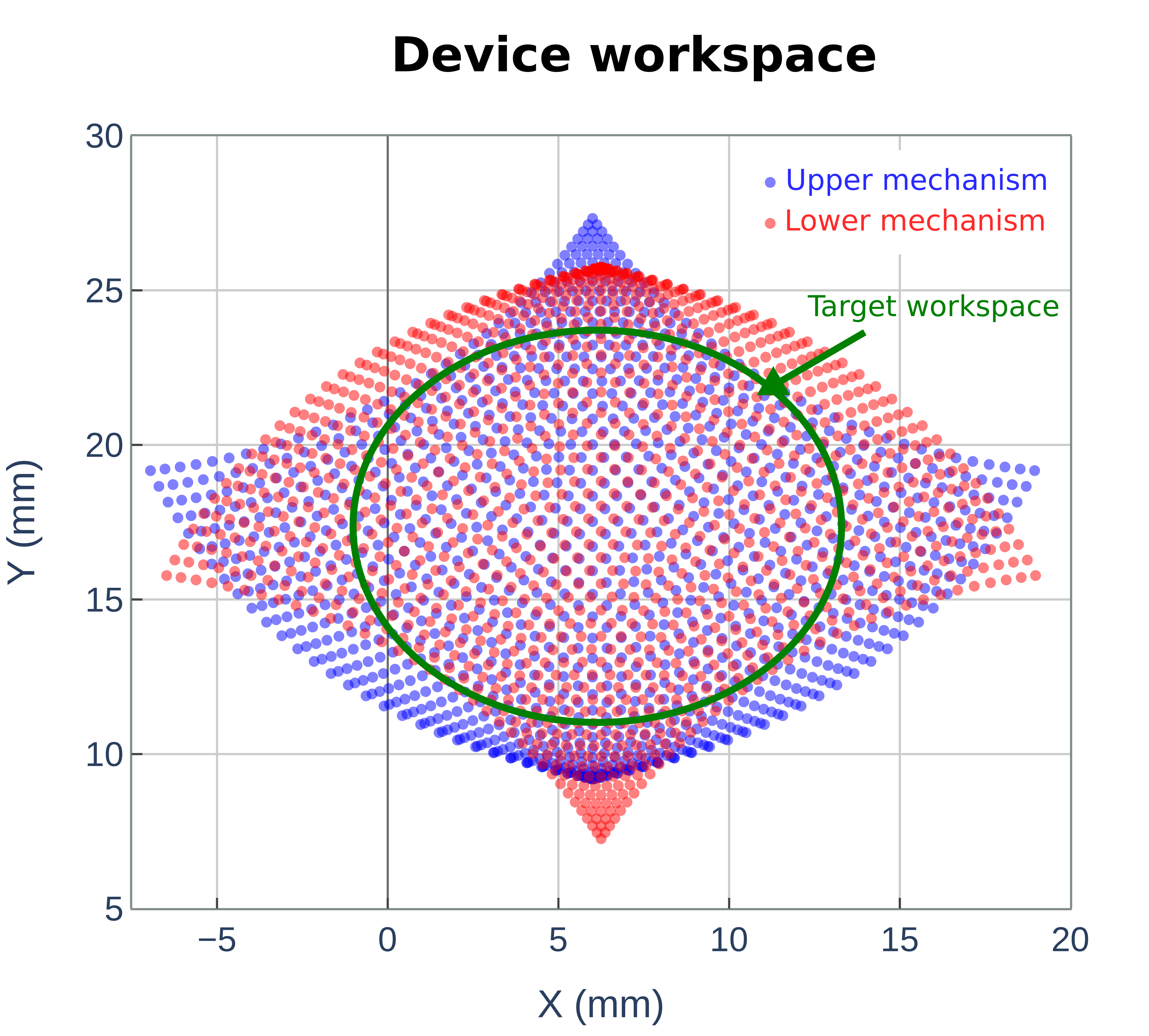}
	\caption{ Device reachable workspace. The blue region corresponds to the reachable workspace for the upper mechanism and the red region to the lower mechanism. The elliptic shape in the center represents the finger tip area of interest.  }
	\label{fig:4}
\end{figure}

The reachable workspace for each device station is different according to the link dimensions. The target workspace for the index finger corresponds to an ellipsoid with dimensions of 15mmx12mm and for the thumb 15mmx14mm. The thumb station reachable and target workspace are depicted in Fig.~\ref{fig:4}. 

\subsection{Device Control}
\begin{figure}[t]
	\centering
	\includegraphics[width=3.4in]{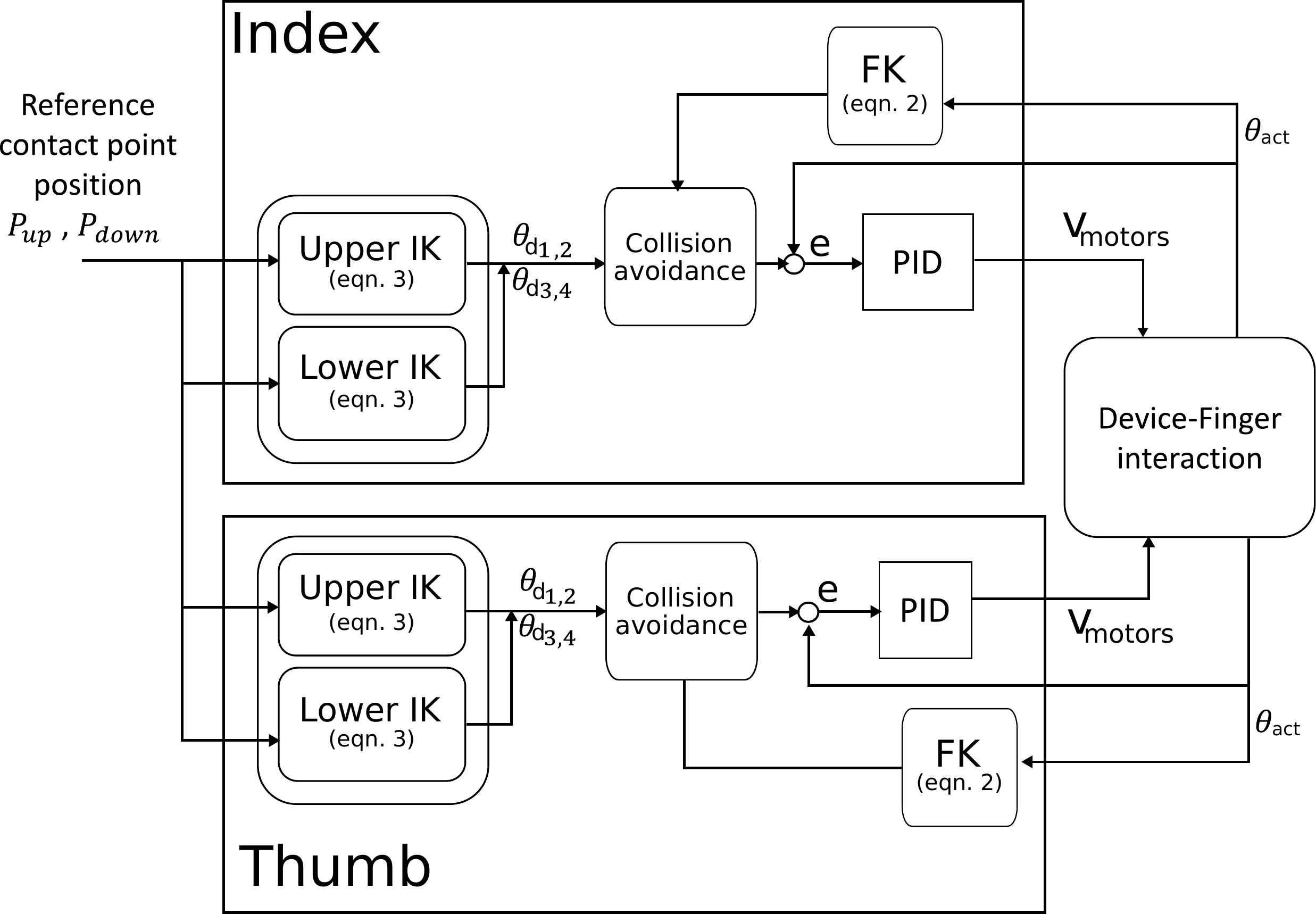}
	\caption{System control architecture. }
	\label{fig:5}
\end{figure}

The motors and sensors are connected to a Raspberry Pi 3 board through a USB data acquisition board USB-1608G (Measurement Computing Co.) and four DC motor driver HATs DFR0592 (DFRobot Inc.). The voltage signals obtained from each potentiometer are acquired by the analog-to-digital converter to estimate the motors joint position. The control signals for each motor are converted to 1kHz PWM signals from the motor drivers. The system control architecture was implemented under the Robot Operating System (ROS) framework and follows the scheme depicted in Fig.~\ref{fig:5}.  The contact point reference positions ($P_{up}$, $P_{down}$) are used as the input for the four Inverse Kinematics (IK) modules of the index finger and thumb. A collision avoidance modules receives the target joint angles (${\theta_{d_{1}}}$, ${\theta_{d_{2}}}$, ${\theta_{d_{3}}}$, ${\theta_{d_{4}}}$) and the current links position estimated with the Forward Kinematics module (FK) using the actual joint angles ($\theta_{act}$). The collision is detected by mesh overlapping. The upper mechanisms are given higher priority, such that in the case of collision the target of the lower mechanism will be replaced by the closest location in the border of the upper mechanism, and the target joint angles will be updated. The motor control signals $v_{motors}$ are generated by a PID controller. The control loop runs at a frequency of 100 Hz. 

\section{Experimental Evaluation and Discussion}
The two experiments are conducted with 5 healthy subjects. 
This study was approved by the Ethics Committee of Nagoya University.

\subsection{Tactile feedback discrimination} 
\begin{figure}[t]
	\centering
	\includegraphics[width=3.25in]{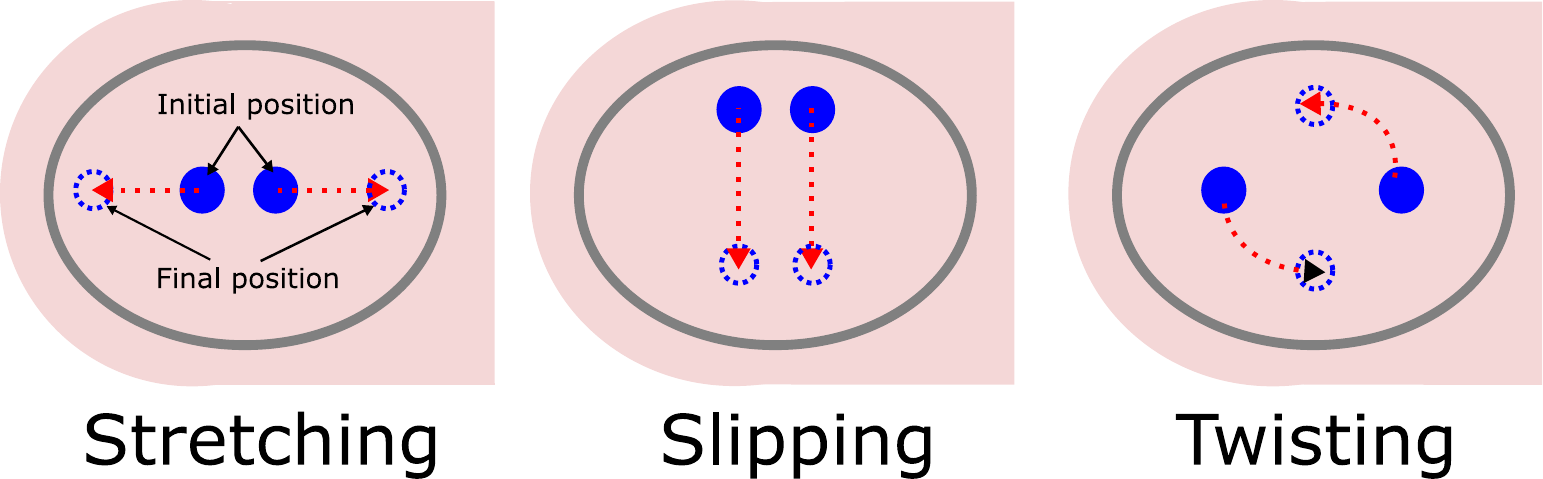}
	\caption{Skin deformation patterns. (a) Stretching. The two tactors are initially placed at the center of the fingertip and follow a horizontal trajectory moving away from the center. (b) Slipping. The tactors start in the upper region of the fingertip and moves towards the lower region of the fingertip. (c) Twisting. The tactors starts separated in the center of the fingertip a moves along a circular path.}
	\label{fig:6}
\end{figure}

\begin{figure}[t]
	\centering
	\includegraphics[width=3.0in]{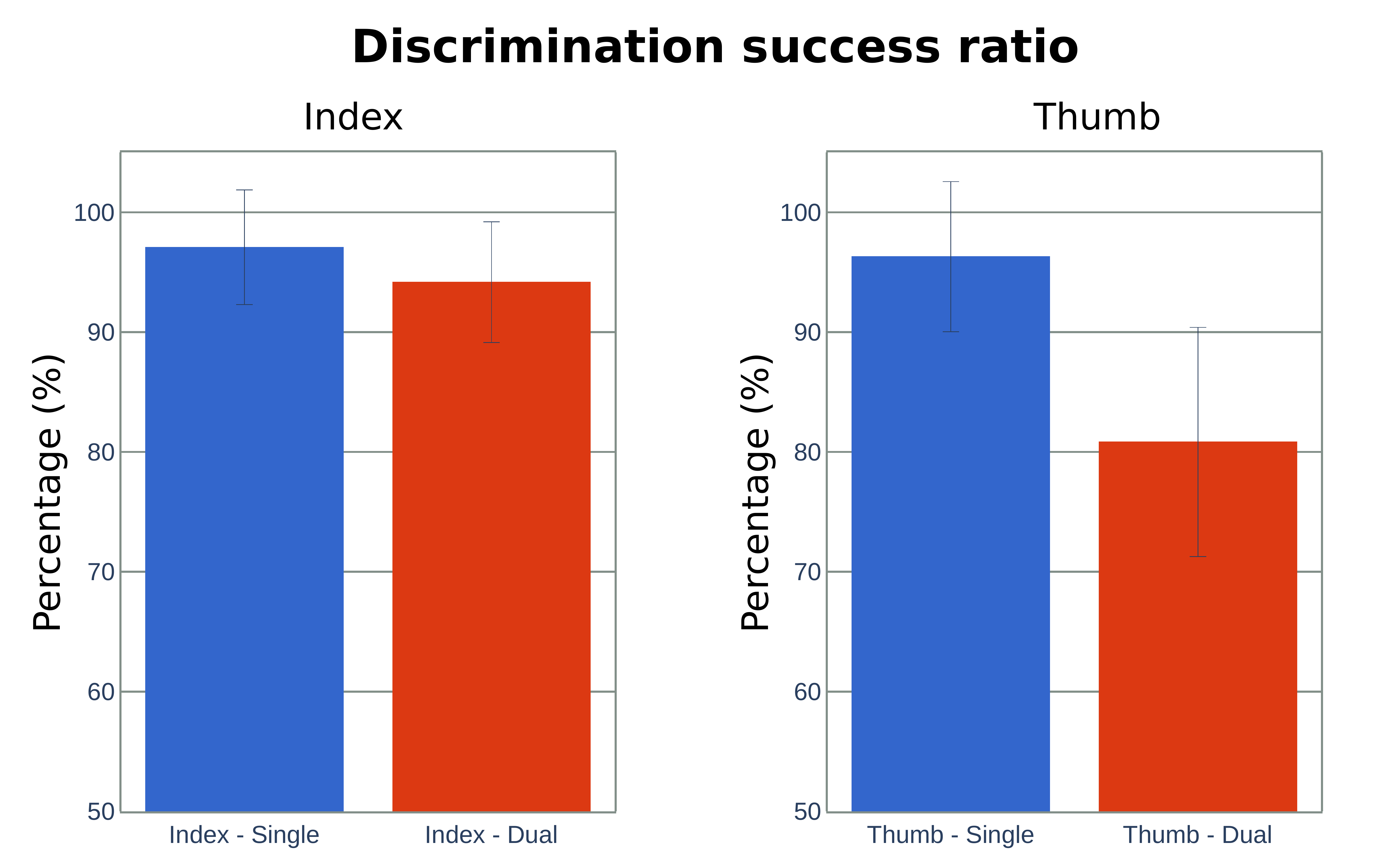}
	\caption{Discrimination success ratio for each finger in Single (only one finger) and Dual (index finger and thumb) modes. (a) Index finger. High success ratio for single and dual modes. (b) Thumb. High success ratio for single but a marked decline is observed for dual mode. }
	\label{fig:7}
\end{figure}

We investigated the user's ability to discriminate shear forces through three skin deformation patterns: skin stretching, skin slipping and skin twisting as shown in Fig.~\ref{fig:6}. The experiment comprises three stages. First, only the index finger is inserted into the device and the subject is asked to identify the pattern between the three modes: stretching, slipping, and twisting. Twenty repetitions are performed selecting randomly the pattern applied. Next, the task is repeated with the thumb. In the last stage, both fingers are inserted in the haptic device and the subject is requested again to identify the skin deformation pattern applied on each finger. This task is repeated 30 times with independent patterns selected randomly for the thumb and index finger. The discrimination success ratio for each finger is depicted in Fig.~\ref{fig:7}. 

The results have shown a high single-finger success discrimination ratio of above 95\% for recognizing between the three deformation patterns. When combining the use of the index finger and thumb, we observe a marked decline of about 15\% for the thumb. This trend is also found when comparing within a single deformation pattern (stretching, slipping or twisting). This could be explained by the conflict existing in discriminating independently when different patterns are applied to each finger. For the index finger, the decline in the ratio is minimum (less than 5\%), so it could be inferred that the user gives a higher priority to the sensation produced in this finger.

\subsection{Teleoperated Pivoting Manipulation}

\subsubsection{Experimental Setup}
The experimental setup of pivoting manipulation is shown in Fig.~\ref{fig:8}. The setup is composed of (1) a LCD screen that provides visual feedback to the subjects, (2) an Omega 7 haptic device to acquire human input to control simulated robot gripper, and provide grasping force feedback, (3) the proposed cutaneous feedback interface attached on the gripper of the omega 7 haptic device, and (4) a Ubuntu desktop PC (Nvidia GPU, RTX 3090) that runs numerical simulator Nvidia Isaac Sim to provide realistic pivoting environment. Data communication between each component is implemented under ROS (Robot Operating System) framework.

\subsubsection{Grasping Force Feedback by Virtual Fixture and Gripper Position Control}
To provide grasping force feedback that indicates the force applied to the object, a haptic virtual fixture is implemented \cite{pruks2020framework}. 
The virtual fixture is implemented as a mass damper system using the following equation,
\begin{equation}
	\begin{aligned}
	    F_{leader} = K (X_{leader} - X_{object}) + D \dot{X}_{leader}
	\end{aligned}
	\label{eq:5}
\end{equation}
where $F_{leader}$ is the force generated by Omega 7 gripper to provide grasping force feedback, $K$ is the desired stiffness of the virtual fixture, and $D$ is the desired damping acting as a stabilization factor, $X_{leader}$ is the gipper distance read from Omega 7 haptic device i.e. desired follower position, and $X_{object}$ is the position in Omega 7 gripper coordinate where the virtual fixture is generated and this position is relevant to object diameter. The gripper inside Nvidia Isaac Sim is controlled by an ideal position controller (PD controller), which adopts a high gain control for positional tracking of $X_{leader}$.

\subsubsection{Experimental Conditions}
\begin{figure*}[t]
	\centering
	\includegraphics[width=0.9\linewidth]{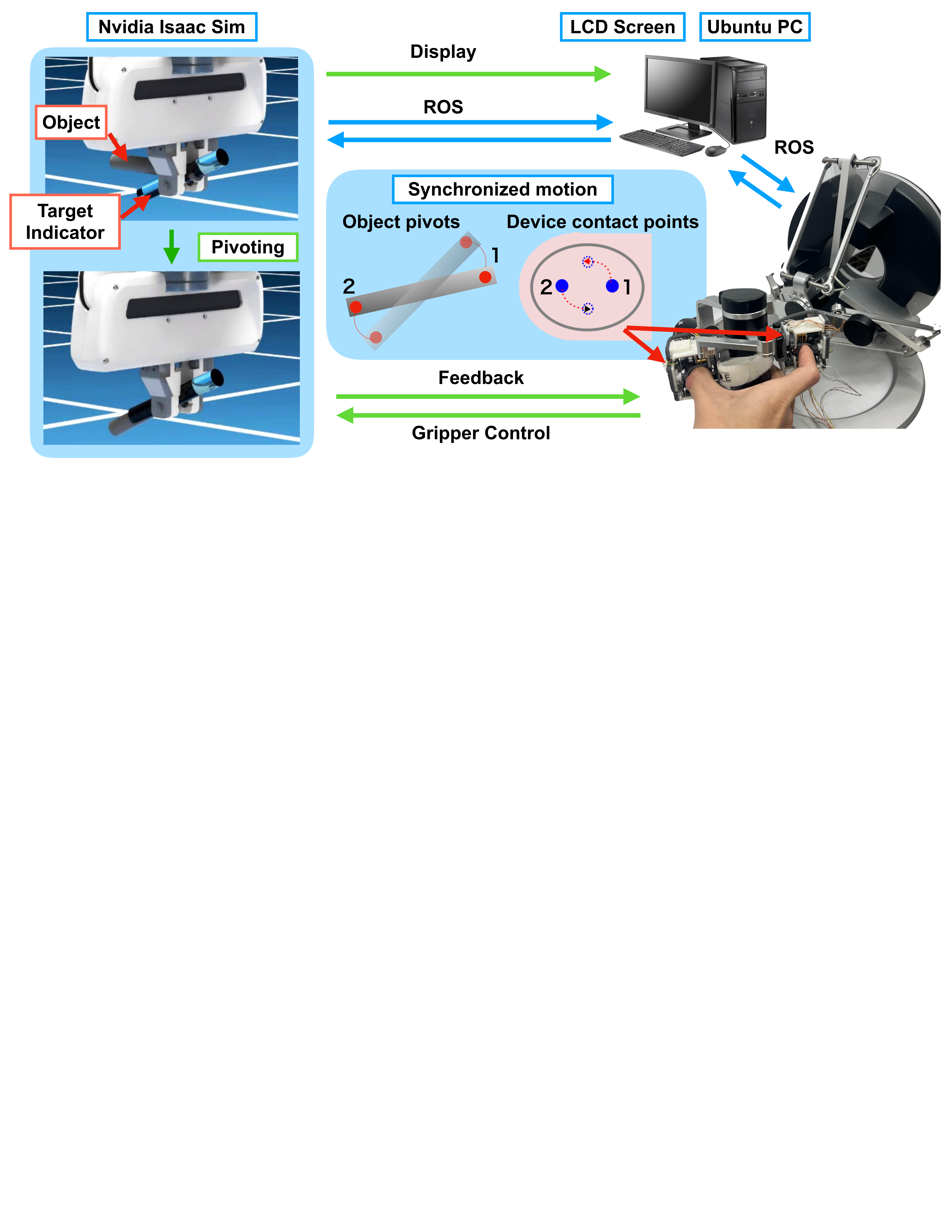}
	\caption{Experimental setup for teleoperated pivoting manipulation }
	\label{fig:8}
\end{figure*}

The object to pivot inside simulator is a rigid body cylinder with a diameter of 1.5 cm and 10 cm in length.
Subjects control the robot finger distance to perform a passive pivoting task, i.e., subjects adjust the grasping force through controlling the finger position to control the object’s rotational motion induced by gravity.  
The pivoting tasks are performed under four distinct conditions. The conditions are as follows:
\begin{enumerate}
  \item Subjects perform the tasks by only using Visual Feedback (\textbf{VF}) from LCD screen. The screen displays Isaac Sim perspective from 45 degree behind the gripper as Fig.~\ref{fig:8} shows. A transparent mesh is loaded to environment for indicating target pivoting angles. 
  \item Subjects perform the tasks by using Visual Feedback and Grasping Force feedback (\textbf{VF + GF}). Grasping force feedback is implemented by referencing to equation (\ref{eq:5}).
  \item Subjects perform the tasks by using Visual Feedback, Grasping Force feedback, and Cutaneous based Tactile Feedback (\textbf{VF + GF + TF}). \textbf{TF} is generated by the cutaneous feedback device, which two contact points tracks the motion of the simulation object in real time to indicate its rotation, i.e. as Fig.~\ref{fig:8} \textit{synchronized motion} shows, in this case the blue dots on the haptic device track the position of the red dots of the object (blue dot 1 tracks red dot 1, and blue dot 2 tracks red dot 2, respectively), following a trajectory represented by the red dotted arrow to generate the cutaneous sensation of twisting.
  \item Subjects perform the tasks by using Visual Feedback and Cutaneous based Tactile Feedback (\textbf{VF + TF}). 
\end{enumerate}

\subsubsection{Experimental Procedures}
Subjects need to perform the pivoting tasks under four conditions. 
Before the experiment, subjects are instructed on the experiment procedure and are given time to practice the operations.
Each condition contains 45 trials, and each trial is composed of random combination of different object masses and target angles. Object's mass and target angles are chosen from 0.005 kg, 0.01 kg, 0.02 kg, 25 degrees, 45 degrees, and 75 degrees, respectively. This random combination simulates the real-world situation where human grasp novel objects, and also prevents subjects from learning the object's characteristics that bias the experiment. In total, the combination yields 9 cases, and each case is guaranteed to be performed for 5 times (9 cases multiplied 5 times yields 45 trials). The completion time of each trial and the angle error between target and actual are recorded automatically. In this experiment the successful task execution is defined as that  angular error is less than $\pm10$ degrees.

\subsubsection{Experimental Results and Discussion}
\begin{figure*}[t]
	\centering
	\includegraphics[width=7in]{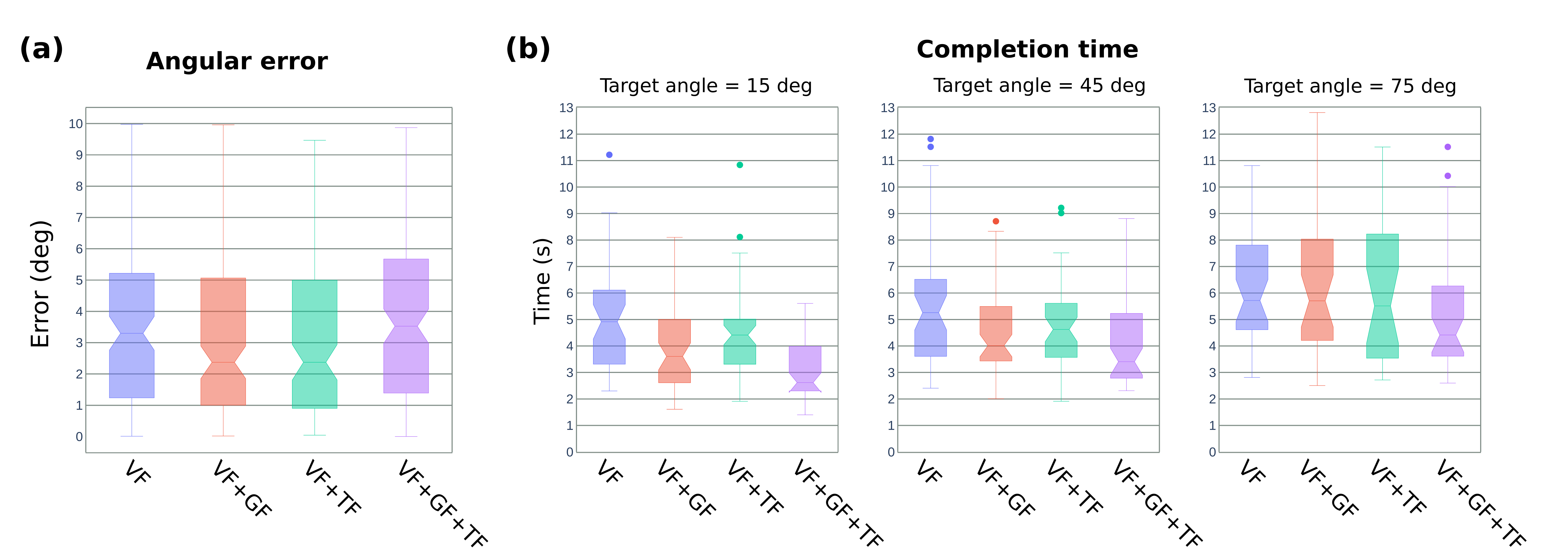}
	\caption{Experiment results of teleoperated pivoting manipulation for each condition. (a) shows angular error between actual object's angle and targeted object's angle. (b) shows the task completion time given different target angles.}
	\label{fig:9}
\end{figure*}

\begin{figure}[t]
	\centering
	\includegraphics[width=3.5in]{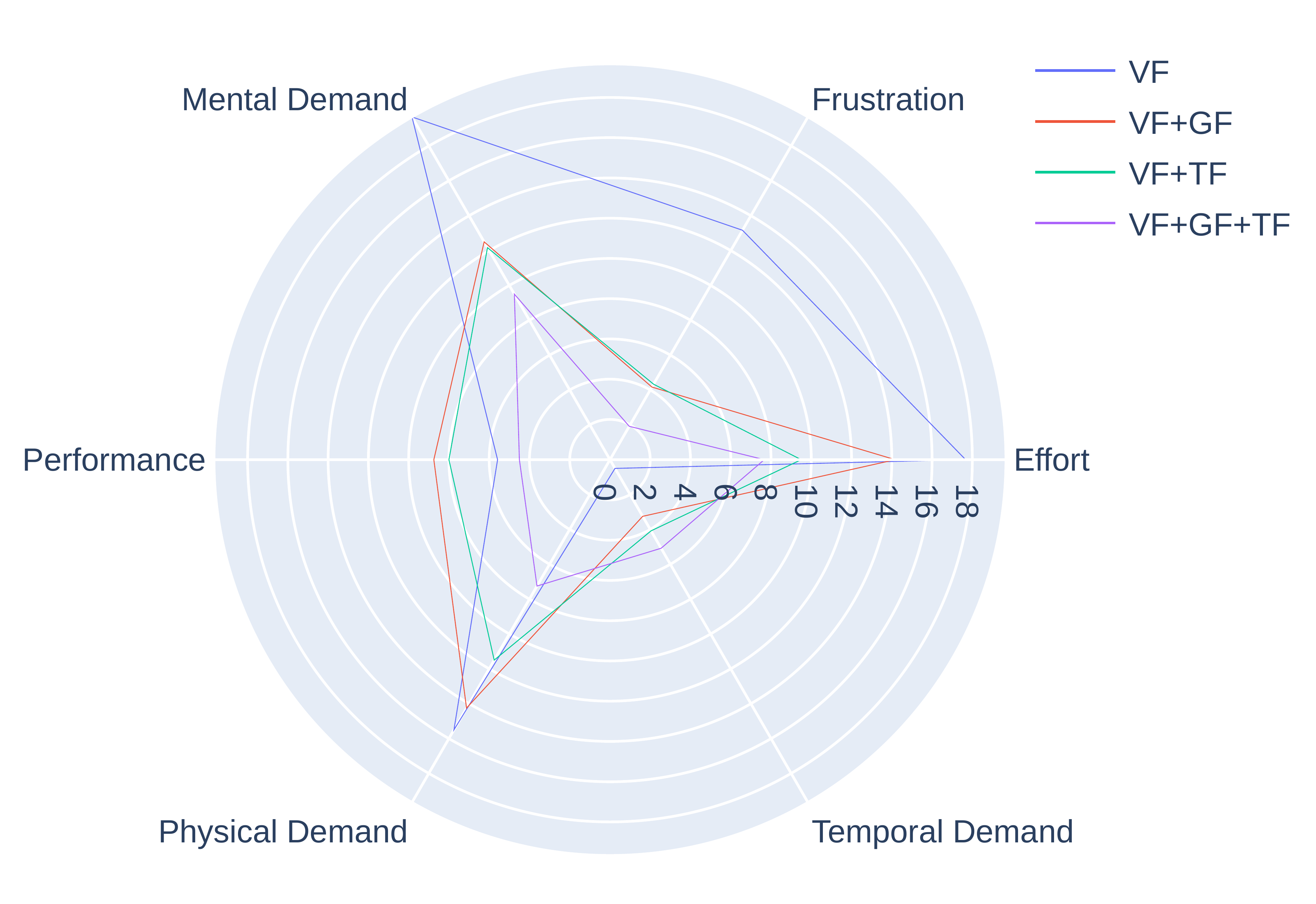}
	\caption{Teleoperated pivoting manipulation: NASA-TLX weighted rating}
	\label{fig:10}
\end{figure}

Figure~\ref{fig:9} \textbf{(a), (b)} shows the experimental results of angular error and task completion time, respectively (N = 5). Figure \ref{fig:10} shows NASA-TLX weighted ratings. The success ratio for each condition, and NASA-TLX overall workload are shown in Table \ref{tab:success_ratio}, and Table \ref{tab:nasa} respectively.

Figure~\ref{fig:9} \textbf{(a)} shows the angular error for each condition (3.29 degree, 2.37 degree, 2.37 degrees, and 5.67 degree for each condition, respectively.) Among each condition, VF+GF+TF produces the most error.
In Figure~\ref{fig:9} \textbf{(b)} mean value of completion time when the target angle equals to 15 degree are 4.9 s, 3.6 s, 4.4 s, and 2.6 s for VF, VF + GF, VF + TF, and VF + GF + TF respectively. When the target angle is 45 degree, the mean values of completion time for the four conditions are 5.2 s, 4.0 s, 4.6 s, and 3.4 s, respectively. And when the target angle is 75 degrees, the mean values of completion time are 5.7 s, 5.7 s, 5.5 s, and 4.4 s respectively. 

The comparison of completion time shows that in condition VF+GF+TF subjects perform pivoting task faster among all conditions. This may imply that the complemented somatosensory feedback (grasping force feedback and shear force feedback) makes subjects more confident when performing pivoting tasks. And this result may be identical to the results reported by neuro-scientists that somatosensory based feedback develops a feedforward model for faster human motion planning \cite{blakemore2002abnormalities}. 
Table \ref{tab:success_ratio} suggests that VF+GF+TF has a close success ratio compared with VF+GF. The result indicates that additional tactile information can help subjects successfully perform in-hand pivoting with improved completion time.
On the contrary, the result of Figure~\ref{fig:9} \textbf{(a)} shows that the condition VF+GF+TF produces the most error. This could be implying that the subjects rely more on haptic feedback and pay less attention to visual feedback for precise adjustment. 

NASA-TLX weighted rating (Figure \ref{fig:10}) shows the least mental demand for condition VF+GF+TF, by means of frustration, effort, and physical demand. While, VF+GF and VF+TF present a similar workload demand, and VF shows the most mental demand, frustration, and effort. This result indicates that, from the subjects point of view, enhanced tactile feedback greatly decreases workload, and the existence of some kind of haptic feedback (GF and TF) also helps to perform tasks with less workload compared with pure visual feedback. The same result is also observed in Table \ref{tab:nasa}, which indicates the existence of haptic feedback decreases the overall workload, and the best performance is produced by complementing the sensory feedback (VF+GF+TF).

\begin{table}[!t]
	\centering
\caption{Success Ratio (\%)}
\begin{tabular}{ |c|c|c|c|  }

 \hline
  VF    & VF+GF  & VF+TF & VF+GF+TF\\
 \hline
 74.4   & 83.3   & 72.4  & 82.7 \\
 \hline
\end{tabular}
\label{tab:success_ratio}
\end{table}

\begin{table}[!t]
	\centering
\caption{NASA-TLX Overall Workload for Each Condition}
\begin{tabular}{ |c|c|c|c|  }
 \hline
  VF   & VF+GF    &   VF+TF & VF+GF+TF \\
 \hline
 71.58  & 57.33 & 49.67 & 36.17 \\
 \hline
\end{tabular}
\label{tab:nasa}
\end{table}

\section{Conclusion and Future Work}

\subsubsection{Conclusion}
This paper proposed a cutaneous feedback interface for in-hand pivoting task.
The aim of this interface is to enhance the haptic feedback needed for successful dexterous teleoperation. The interface complements the haptic sensation (shear force information, grasping force information ) humans rely on when performing dexterous skills. 
The interface is designed based on five-bar link mechanisms. The novelty of the interface is to provide multi-point contacts for cutaneous feedback.

A passive pivoting task inside a numerical simulator Isaac Sim is conducted to verify the effect of proposed cutaneous feedback interface. The performance of the interface is analysed and verified by conducting experiments. In the skin deformation pattern recognition experiment, the results have shown a high single finger success discrimination
ratio of above 95\% on recognizing between the three deformation patterns. While combining the use of the index finger and the thumb, a marked decline of about 15\% for the thumb is observed.
In the teleoperated pivoting experiment, the results have shown the fastest task completion time for the condition that combines cutaneous feedback and grasping force feedback. And NASA-TLX weighted rating indicated a decrease in the mental workload when using both cutaneous feedback and grasping force feedback. 
The present work explores the potential use of cutaneous information to substitute conventional force feedback, and the benefits of complemented sensory perception for dexterous teleoperated in-hand manipulation.

\subsubsection{Future Work}
Future work will be conducting experiments with more subjects and expanding the research to real environments with real robot grippers. In a real-world environment, it is hard to acquire fast and accurate feedback of object's orientation, hence integrating the cutaneous feedback interface with tactile sensors attached on gripper's finger tip will be carried out.

\bibliographystyle{IEEEtran}
\bibliography{iros}

\end{document}